\PassOptionsToPackage{hidelinks}{hyperref}
\documentclass[lettersize,journal]{IEEEtran}
\usepackage{amsmath,amsfonts}
\usepackage{algorithmic}
\usepackage{algorithm}
\usepackage{array}
\usepackage[caption=false,font=normalsize,labelfont=sf,textfont=sf]{subfig}
\usepackage{textcomp}
\usepackage{stfloats}
\usepackage{url}
\usepackage{verbatim}
\usepackage{graphicx}
\usepackage{cite}
\usepackage{multirow}
\usepackage{xcolor}
\usepackage{orcidlink}
\usepackage{hyperref}
\usepackage{xurl}
\begin{document}
\title{GAViD: A Large-Scale Multimodal Dataset for Context-Aware Group Affect Recognition from Videos}
\author{
Deepak~Kumar\orcidlink{0000-0002-8622-7251}\thanks{Corresponding author: Deepak Kumar (email: d\_kumar@cs.iitr.ac.in)},
Abhishek~Pratap~Singh\orcidlink{0009-0003-7593-2347},
Puneet~Kumar\orcidlink{0000-0002-4318-1353},~\IEEEmembership{Member,~IEEE,}
Xiaobai~Li\orcidlink{0000-0003-4519-7823},~\IEEEmembership{Senior~Member,~IEEE,}
and~Balasubramanian~Raman\orcidlink{0000-0001-6277-6267},~\IEEEmembership{Senior~Member,~IEEE}%
\thanks{
Deepak Kumar, Abhishek Pratap Singh, and Balasubramanian Raman are with the Department of Computer Science and Engineering, Indian Institute of Technology Roorkee, Roorkee, India.}
\thanks{
Puneet Kumar is with the Indian Institute of Technology Ropar, Punjab, India.}
\thanks{
Xiaobai Li is with the Zhejiang University, China and University of Oulu, Finland.}
}

%\thanks{Manuscript received April 19, 2021; revised August 16, 2021.}}

\markboth{Journal of \LaTeX\ Class Files,~Vol.~14, No.~8, August~2021}%
{Shell \MakeLowercase{\textit{et al.}}: A Sample Article Using IEEEtran.cls for IEEE Journals}

\IEEEpubid{0000--0000/00\$00.00~\copyright~2021 IEEE}

\maketitle

\begin{abstract}
%Recognizing group affect \emph{in-the-wild} settings remains challenging due to two key factors: the difficulty of capturing and labeling group data and the complexity of analyzing group affect amid diverse interactions and contextual variability. The lack of comprehensive datasets annotated with multimodal and contextual information further limits advances in the field. To address this, we introduce the Group Affect from ViDeos (GAViD) dataset, comprising 5091 video clips with multimodal data (video, audio and context), annotated with ternary valence and discrete emotion labels and enriched with VideoGPT-generated contextual metadata and human-annotated action cues. We also present Context-Aware Group Affect Recognition Network (CAGNet) for multimodal context-aware group affect recognition. CAGNet achieves 61.20\% test accuracy on GAViD, comparable to state-of-the-art performance. The dataset and code are available at \href{https://github.com/deepakkumar-iitr/GAViD}{\underline{github.com/deepakkumar-iitr/GAViD}}.
Understanding affective dynamics in real-world social systems is fundamental to modeling and analyzing human-human interactions in complex environments. Group affect emerges from intertwined human–human interactions, contextual influences, and behavioral cues, making its quantitative modeling a challenging computational social systems problem. However, computational modeling of group affect in \emph{in-the-wild} scenarios remains challenging due to limited large-scale annotated datasets and the inherent complexity of multimodal social interactions shaped by contextual and behavioral variability. The lack of comprehensive datasets annotated with multimodal and contextual information further limits advances in the field. To address this, we introduce the Group Affect from ViDeos (GAViD) dataset, comprising 5091 video clips with multimodal data (video, audio and context), annotated with ternary valence and discrete emotion labels and enriched with VideoGPT-generated contextual metadata and human-annotated action cues. We also present Context-Aware Group Affect Recognition Network (CAGNet) for multimodal context-aware group affect recognition. CAGNet achieves 63.20\% test accuracy on GAViD, comparable to state-of-the-art performance. The dataset and code are available at \href{https://github.com/deepakkumar-iitr/GAViD}{\underline{github.com/deepakkumar-iitr/GAViD}}.
%The data, code and model will be available upon the paper's acceptance (currently not shared for double-blind anonymity). 
\end{abstract}


\section{Introduction}
\IEEEPARstart{G}{roup} Affect Recognition (GAR) is a subfield of affective computing focused on analysing emotion patterns across individuals to reveal collective affect~\cite{petrova2020group}. Traditionally, affective computing has emphasised individual-level emotion recognition~\cite{khare2024emotion,guo2024development}, with far less attention given to group-level, intersubjective analysis~\cite{dhall2020emotiw,petrova2020group}. Yet, GAR is gaining momentum with applications in business productivity, marketing, team performance evaluation, social psychology and human-computer interaction \cite{otamendi2020emotional, quiroz2022group, huang2024survey}. It also aids entertainment, public safety and understanding group emotions in education \cite{pervez2024affective, kalyta2023facial, kumar2024measuring}.

GAR in real-world uncontrolled environments, also called \emph{in-the-wild} settings, faces two broad categories of challenges. The first concerns the difficulty of capturing group scenes and labelling them. Videos often suffer from poor lighting, variable visibility, motion blur, occlusion, changing camera perspectives, and temporal dynamics that complicate tracking individuals across frames~\cite{augusma2023multimodal, sun2020multi}. Additionally, there is a lack of large, well-annotated datasets with diverse group-level labels, such as sentiment and discrete emotion categories, which are essential for developing robust models but remain largely unavailable~\cite{petrova2020group, dhall2020emotiw, sharma2021audio}. The second category of challenges is related to analysis and recognition: group affect is complex due to dynamic and nuanced group interactions, diverse scene contexts, and the interplay of individual emotions within a group~\cite{kosti2017emotion,malhotra2023social}.

A variety of GAR techniques have been proposed to address these challenges, leveraging multimodal fusion, graph neural networks, attention mechanisms, multi-task learning, temporal modelling, activity analysis, transfer learning and domain adaptation~\cite{sun2020multi, pinto2020audiovisual, guo2020graph, wang2020implicit, li2022dynamic}. However, the lack of comprehensive datasets with multimodal and contextual annotations further limits progress in the field~\cite{petrova2020group, dhall2020emotiw}. Additionally, most GAR methods rely on visual features that seldom capture the full context needed for robust analysis \cite{sun2020multi, sharma2021audio}. Recent advances in Large Language Models (LLMs) demonstrate that contextual cues can significantly benefit video captioning and narrative understanding~\cite{chang2024survey}. Yet, contextual modelling remains underexplored in GAR, mainly due to limited or context-poor datasets.

%placed here to make it appear on page 2, top
\begin{table*}[ht]
\centering
\renewcommand{\arraystretch}{1.3} % Increase row height (try 1.2–1.5)
\caption{Summary of existing GAR datasets. Here, `Valence Labels' refer to Positive (P), Negative (N) and Neutral (Ne) group-level labels; `Emotion Labels' refer to discrete emotion classes such as happy, sad, fear, anger, etc; $\checkmark$: available, $\times$: not available.}
\label{tab:datasets}
\resizebox{.93\textwidth}{!}{
\begin{tabular}{|l|c|c|c|c|c|}
\hline
\multicolumn{6}{|c|}{\textbf{Image-based GAR Datasets}} \\
\hline
\textbf{Dataset} & \textbf{Year} & \# \textbf{Images} & \textbf{Valence Labels} & \textbf{Emotion Labels} & \textbf{Contextual Info} \\
\hline
FindingEmo~\cite{mertens2024findingemo} & 2024 & 25000 & P, N, Ne & 8 classes     & $\times$ \\\hline
%EmoSet~\cite{yang2023emoset}  & 2023 & 60\,000 & P, N     & 8 classes               & $\checkmark$ \\\hline
GroupEmoW~\cite{guo2020graph}          & 2020 & 15894 & P, N, Ne & $\times$                & $\times$ \\\hline
GAFF 3.0~\cite{dhall2018emotiw}        & 2018 & 17172 & P, N, Ne & $\times$                & $\times$ \\\hline
GAFF 2.0~\cite{dhall2017individual}    & 2017 & 6471  & P, N, Ne & $\times$                & $\times$ \\\hline
EmotiC~\cite{kosti2017emotion}         & 2017 & 18316 & P, N, Ne & 26 classes              & $\times$ \\\hline
HAPPEI~\cite{dhall2015automatic}       & 2015 & 4886  & $\times$ & Only `Happy' (6 levels) & $\times$ \\\hline
MultiEmoVA~\cite{mou2015group}         & 2015 & 250     & P, N, Ne & $\times$ (Arousal only) & $\times$ \\\hline
\multicolumn{6}{|c|}{\textbf{Video-based GAR Datasets}} \\ % small vertical gap
\hline
\textbf{Dataset} & \textbf{Year} & \# \textbf{Videos/Clips} & \textbf{Valence Labels} & \textbf{Emotion Labels} & \textbf{Contextual Info} \\\hline
OUC-CGE~\cite{lu2025video}    & 2025 & 350 Clips    & P, N, Ne & $\times$ & $\checkmark$ \\\hline
VEATIC~\cite{ren2024veatic}   & 2024 & 124 Clips    & P, N, Ne & $\times$ & $\checkmark$ \\\hline
G-REx~\cite{bota2024real}     & 2024 & 380 Sessions & P, N, Ne & $\times$ & $\checkmark$ \\\hline
GCE~\cite{gce}                         & 2024 & 1029 Clips & P, N, Ne & $\times$ & $\times$ \\\hline
GECV~\cite{gecv}                       & 2022 & 627 Videos   & P, N, Ne & $\times$ & $\times$ \\\hline
VGAF~\cite{sharma2021audio}            & 2021 & 4183 Clips & P, N, Ne & $\times$ & $\times$ \\\hline
AMIGOS~\cite{miranda2018amigos}        & 2018 & 80 Videos    & $\times$ (individual-level $\checkmark$) & $\times$ (individual-level $\checkmark$) & $\checkmark$ \\\hline
MatchNMingle~\cite{cabrera2018matchnmingle} & 2018 & 674 Videos & $\times$ & $\times$ & $\checkmark$ \\\hline
MED~\cite{rabiee2016novel}             & 2016 & 31 Videos    & $\times$ & $\times$ & $\checkmark$ \\\hline
SALSA~\cite{alameda2015salsa}          & 2015 & 18 Videos    & $\times$ & $\times$ & $\checkmark$ \\\hline
Violent Flows~\cite{hassner2012violent}& 2012 & 246 Videos   & $\times$ & $\times$ & $\checkmark$ \\\hline
PETS~\cite{pets2009overview}           & 2009 & 59 Videos    & $\times$ & $\times$ & $\checkmark$ \\\hline
\end{tabular}
}
\end{table*}

To bridge these gaps, we introduce GAViD, a large-scale, \emph{in-the-wild} multimodal dataset with 5091 samples containing video clips, audio, and contextual information, along with ternary valence and discrete emotion annotations. It is enriched with contextual metadata generated by multimodal LLMs (MLLMs)~\cite{kasneci2023chatgpt} and human-labelled action cues, emotion intensity levels, and interaction type information. We also propose Context-Aware Group Affect Recognition Network (CAGNet), which achieves 63.20\% valence classification accuracy on GAViD, comparable to the state-of-the-art. This paper presents results for valence classification and discrete emotion recognition. Our main contributions are summarised as follows. 

\vspace{1 mm}%-.03in}
\begin{itemize} %[leftmargin=*]
\item We introduce GAViD, a large-scale \emph{in-the-wild} dataset for GAR, containing multimodal data (video, audio and context) and annotated with ternary valence and discrete emotion labels.
\enlargethispage{-\baselineskip}
\item 
GAViD is enriched with contextual metadata generated via multimodal LLMs and action cues through a human annotation campaign with over 108 annotators, ensuring reliable labels.
%\enlargethispage{-0.7\baselineskip}

\item We present CAGNet, a GAR model that fuses visual, audio and context features for group affect recognition.
\end{itemize}

%-*-*-*-*-*-              Literature Review -*-*-*-*-*-
%====================================
\section{Related Works}\label{sec:lr} 
To position our contributions, we review related works, summarise existing GAR datasets (Table~\ref{tab:datasets}) and highlight the gaps we address.

\subsection{Group Affect Recognition}
Most GAR methods used supervised learning on annotated audio–visual clips, training multimodal models to predict group affect~\cite{slogrove2022group,augusma2023multimodal}. For instance, Zhu et al.~\cite{zhu2025towards} proposed an uncertainty-aware framework for robust representation learning, while Sharma et al.~\cite{sharma2021audio} introduced VGAFNet for audio–visual affect prediction. Malhotra et al.~\cite{malhotra2023social} employed multi-task learning to jointly predict social events and group affect, whereas Augusma et al.~\cite{augusma2023multimodal} focused on privacy-compliant recognition using only global features. Slogrove et al.~\cite{slogrove2022group} investigated pose cues, while others~\cite{sun2020multi,liu2020group} fused facial and scene-level features to boost accuracy. Guo et al.~\cite{guo2018group} introduces a hybrid deep learning model that fuses face, scene, skeleton, and attention-based visual features for group-level emotion recognition. Zhang et al.~\cite{zhang2021real} proposes ERLDK, a reinforcement learning–based multimodal model that enables real-time conversational emotion recognition by leveraging emotion-pair context, domain knowledge, and a DDQN–GRU framework. Recently, Huang et al.~\cite{huang2025multimodal} proposes MSST, a multimodal spatiotemporal semi-supervised transformer that effectively fuses spatial, temporal, and multimodal cues for group-level emotion recognition.
%{\color{blue}\textbf{Add this \cite{huang2025multimodal} new paper at appropriate place}}

\subsection{Group Context Understanding}
Existing GAR studies have leveraged global scene context or the presence of detected objects to infer group emotions~\cite{huang2019analyzing, fujii2020hierarchical, gecv}. Building on this, several complementary approaches enrich emotion representations by incorporating individual-level object features or modelling social interactions within the group~\cite{zhu2025towards, wang2022congnn, fujii2020hierarchical}. In addition, both non-verbal contextual cues such as body posture~\cite{slogrove2022group} and facial expressions~\cite{sharma2021audio} and verbal cues~\cite{kumar2025fusing} have been shown to play significant roles. Moreover, contextual information such as the event details and background objects further provides useful context that influences group emotion recognition~\cite{kosti2017emotion,malhotra2023social}.

% Placed here for better formatting
%\setcounter{table}{2}
\begin{table*}[!ht]
\centering	
\caption{Samples from the GAViD dataset illustrating input modalities and valence label computation. VA\_Prob, VC\_Prob and AC\_Prob denote the output probabilities from the Visual–Audio, Visual–Context and Audio–Context cross‐attention blocks. Final\_Prob aggregates these probabilities to determine the predicted valence label. The highest probability values are marked in bold. \vspace{-.05in}} %Discrete emotion labels are determined using an identical fusion procedure.
\label{tab:sample_data}
\resizebox{.98\textwidth}{!}{%
    \begin{tabular}{l}
    \includegraphics[width=1\textwidth]{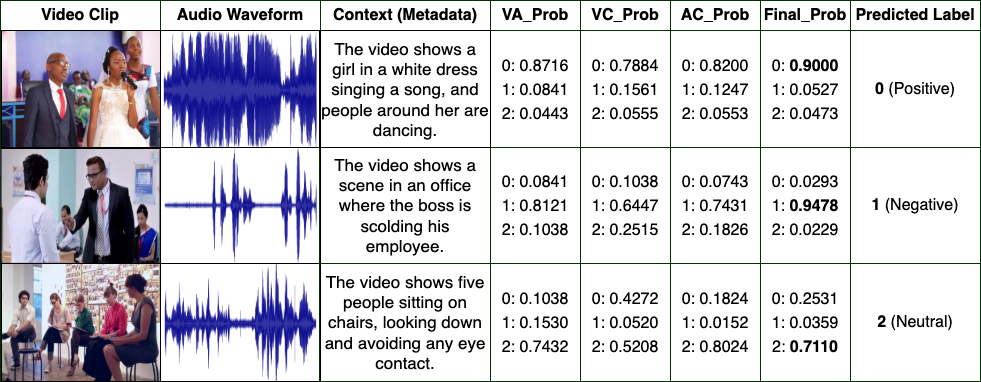}
    \end{tabular}%
}
\end{table*}

\subsection{Existing GAR Datasets} %Addressing the group affect recognition (GAR) problem requires both robust algorithms and high-quality data, as algorithmic performance heavily depends on the richness of the dataset. 
Several vision-based datasets have been developed to support GAR research, focusing on either static images or videos containing multiple individuals (Table~\ref{tab:datasets}). Image-based datasets such as FindingEmo~\cite{mertens2024findingemo}, HAPPEI~\cite{dhall2015automatic}, MultiEmoVA~\cite{mou2015group}, GAFF 2.0~\cite{dhall2017individual}, GAFF 3.0~\cite{dhall2018emotiw}, EmotiC~\cite{kosti2017emotion} and GroupEmoW~\cite{guo2020graph} primarily use static frames sourced from the web, typically containing scenes with more than two people. These datasets generally offer coarse emotion labels (e.g., positive, neutral, negative) or valence ratings, with some providing finer details. For example, HAPPEI annotates six levels of group happiness at individual and group levels, while MultiEmoVA includes continuous valence and arousal scores. GAF, with 17,172 images, is the largest in scale. However, these datasets largely lack contextual or scene-level metadata.

Video-based datasets introduce temporal dynamics but still have notable gaps. GCE~\cite{gce} includes 1,029 segmented YouTube videos featuring various interactions such as interviews, meetings, and informal discussions. GECV~\cite{gecv} contains 627 videos captured in leisure and crowded settings, while VGAF~\cite{sharma2021audio} offers 4,183 samples of group interactions from YouTube. All three focus only on ternary valence. AMIGOS~\cite{miranda2018amigos} and MatchNMingle~\cite{cabrera2018matchnmingle} incorporate self-reports and social cues but lack group-level emotion annotations. Other datasets, such as OUC-CGE~\cite{lu2025video}, VEATIC~\cite{ren2024veatic}, and G-REx~\cite{bota2024real}, provide contextual information but omit discrete emotions. Crowd-focused datasets (MED~\cite{rabiee2016novel}, Violent Flows~\cite{hassner2012violent}, PETS~\cite{pets2009overview}) emphasise behavioural tags (e.g., fighting, panic, boundary crossing) without affective labels.

Overall, no existing dataset offers a comprehensive combination of valence, discrete emotions, and contextual metadata, highlighting the need for a multimodal, context-rich GAR dataset. To bridge this gap, we introduce GAViD, a large-scale \emph{in-the-wild} dataset of 5091 clips with aligned video, audio and contextual information. Unlike existing datasets, which offer at most two of the three annotation types (valence, discrete emotion labels and contextual metadata), GAViD provides all three. Contextual metadata about scene type and event details is generated by multimodal LLMs~\cite{kasneci2023chatgpt} and verified by human annotators. The annotators also label action cues, emotion intensity levels and interaction types (cooperative, hostile and neutral), delivering a comprehensive affect profile. %We further propose CAGNet, a baseline multimodal context-aware network that fuses visual, audio and contextual features, demonstrating the value of our enriched annotations for robust, interpretable real-world GAR applications.

%%%%%%%%%%%%%%%%%%%%%%%%%%%%%%%%%%%%%%%%%%
%====================================
%-*-*-*-*-*- Dataset Description -*-*-*-*-*-
%====================================
\section{GAViD Dataset}\label{sec:data} 
GAViD is a large-scale, \emph{in-the-wild} multimodal dataset of 5091 samples, each annotated with the elements listed below. The following sections describe its key details and compilation procedure. 
% \vspace{-.07in}
\begin{enumerate} %[leftmargin=*, label=\arabic*)]
  \item Raw video clips of an average duration of five seconds,
  \item Audio aligned with the video clips,
  \item Contextual metadata (scene descriptions, event labels) generated by a multimodal LLM and human-verified,
  \item Group affect labels: ternary valence (positive, neutral, negative) and five discrete emotions (happy, sad, fear, anger, neutral),
  \item Emotion intensity ratings (high, medium, low),
  \item Interaction type labels (cooperative, hostile, neutral),
  \item Action cues (e.g.\ smiling, clapping, shouting, dancing, singing).
\end{enumerate}

%---
\subsection{Dataset Compilation Process} %Construction/compilation.
\subsubsection{Data Collection}
\begin{figure*}[!t]
\centering
\includegraphics[width=0.98\linewidth]{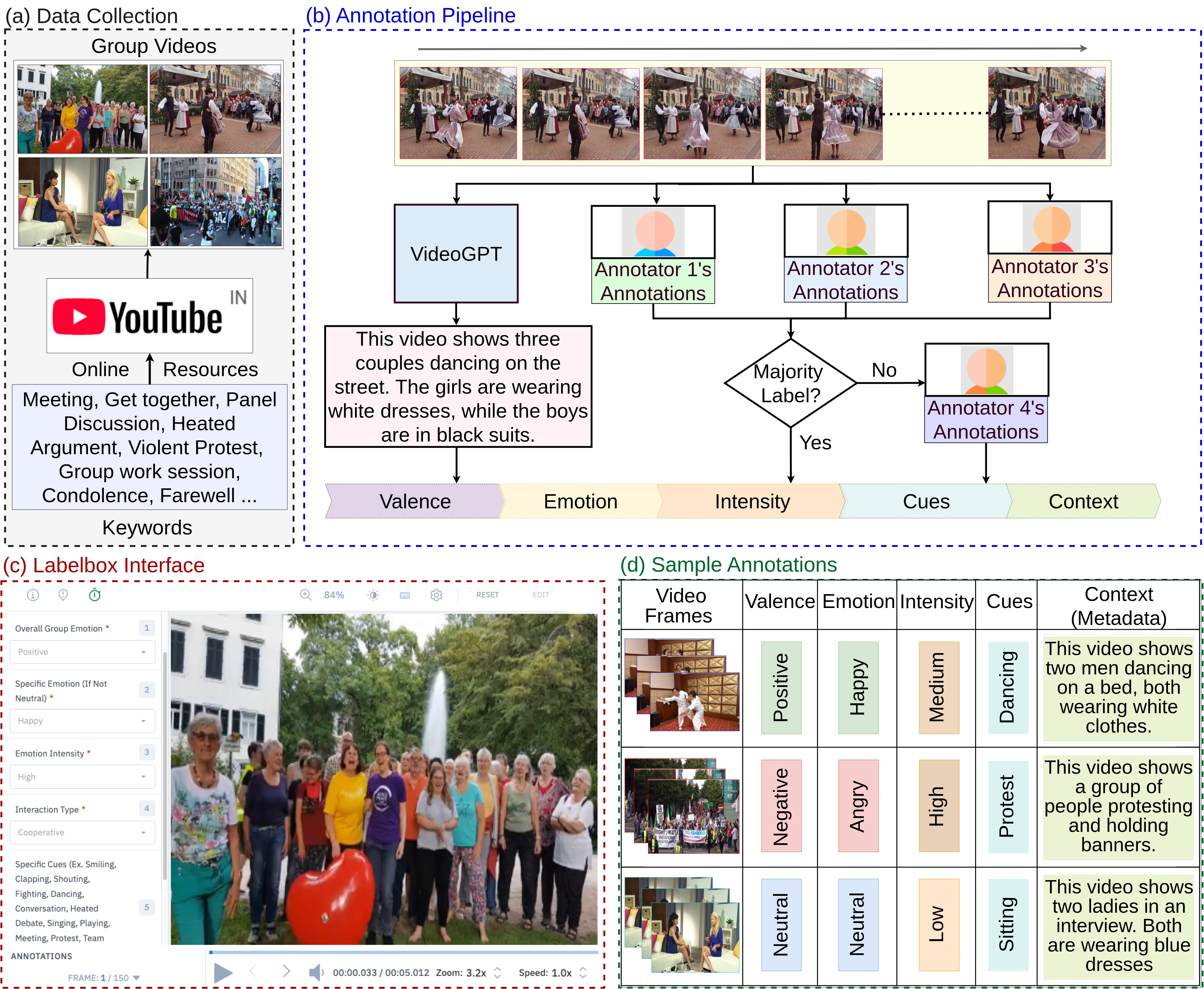} 
\caption{Overview of the GAViD annotation pipeline and interface. The diagram illustrates stages of data collection, sample video frames with valence, emotion, intensity, cues and contextual labels, and the Labelbox interface used for multi-annotator input. Sample context descriptions and VideoGPT-suggested keywords demonstrate how human and AI annotations are integrated.}
\label{fig:data} 
\end{figure*}

As depicted in Fig.~\ref{fig:data}(a), videos were sourced from YouTube under the Creative Commons \emph{CC BY} license, following the ethical protocols mentioned in Section~\ref{sec:ethics}. To capture diverse affective states, we used keyword-based queries, which will be shared in the GitHub repository after the paper's acceptance (currently not shared for double-blind anonymity). %in the \href{https://zenodo.org/records/15448846}{\underline{GAViD repository (zenodo.org/records/15448846)}} 
We included only videos that were CC-licensed and featured two or more individuals to capture real-world group dynamics. Applying these criteria yielded 321 raw videos for our source corpus. As shown in Fig.~\ref{fig:data}(b), VideoGPT MLLM generates a scene's contextual metadata that annotators verify in further phases. 

\begin{figure*}[!htp]
\centering
\includegraphics[width=65mm,height=45mm]{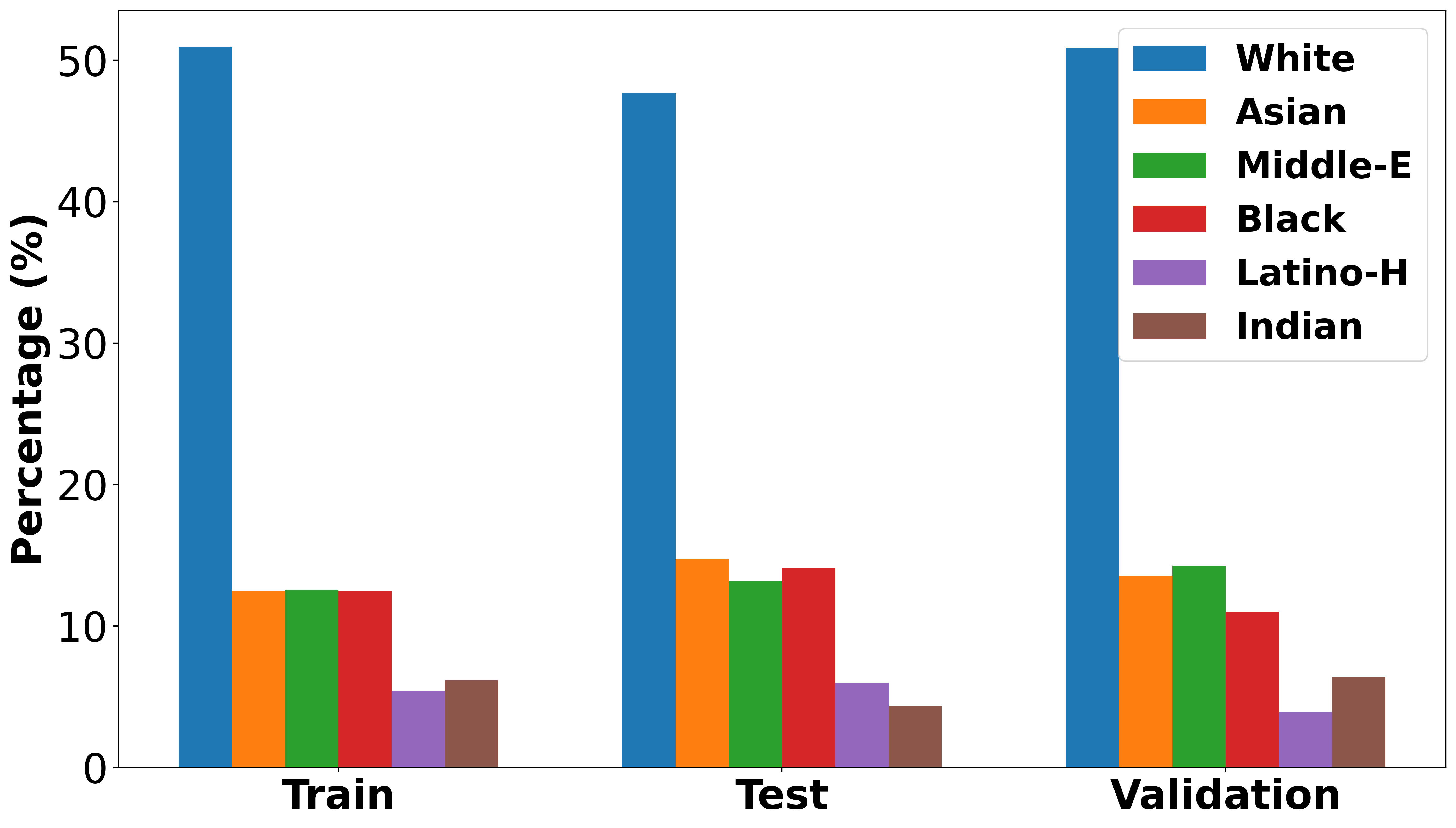}
\includegraphics[width=45mm,height=45mm]{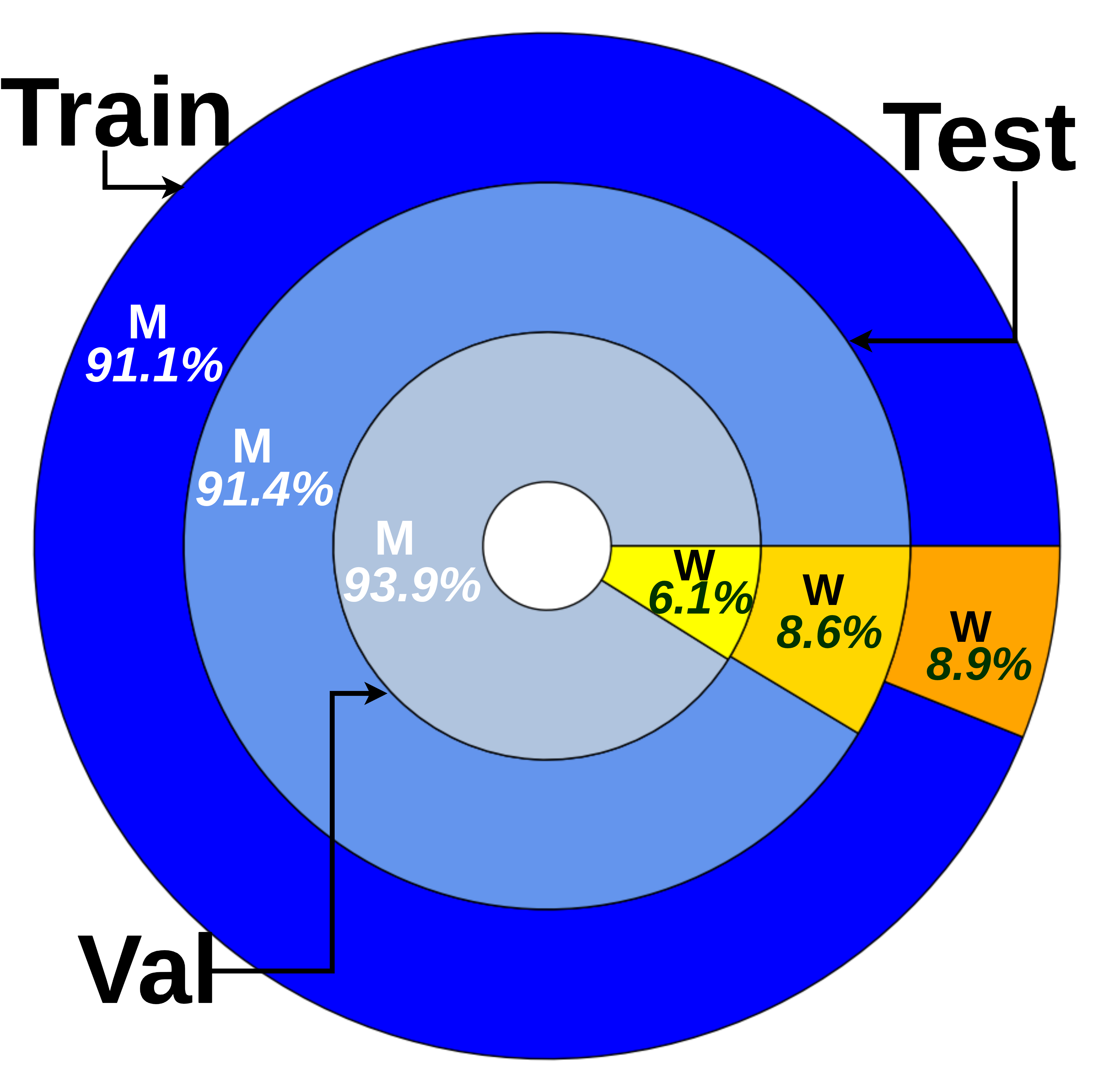}
\includegraphics[width=65mm,height=45mm]{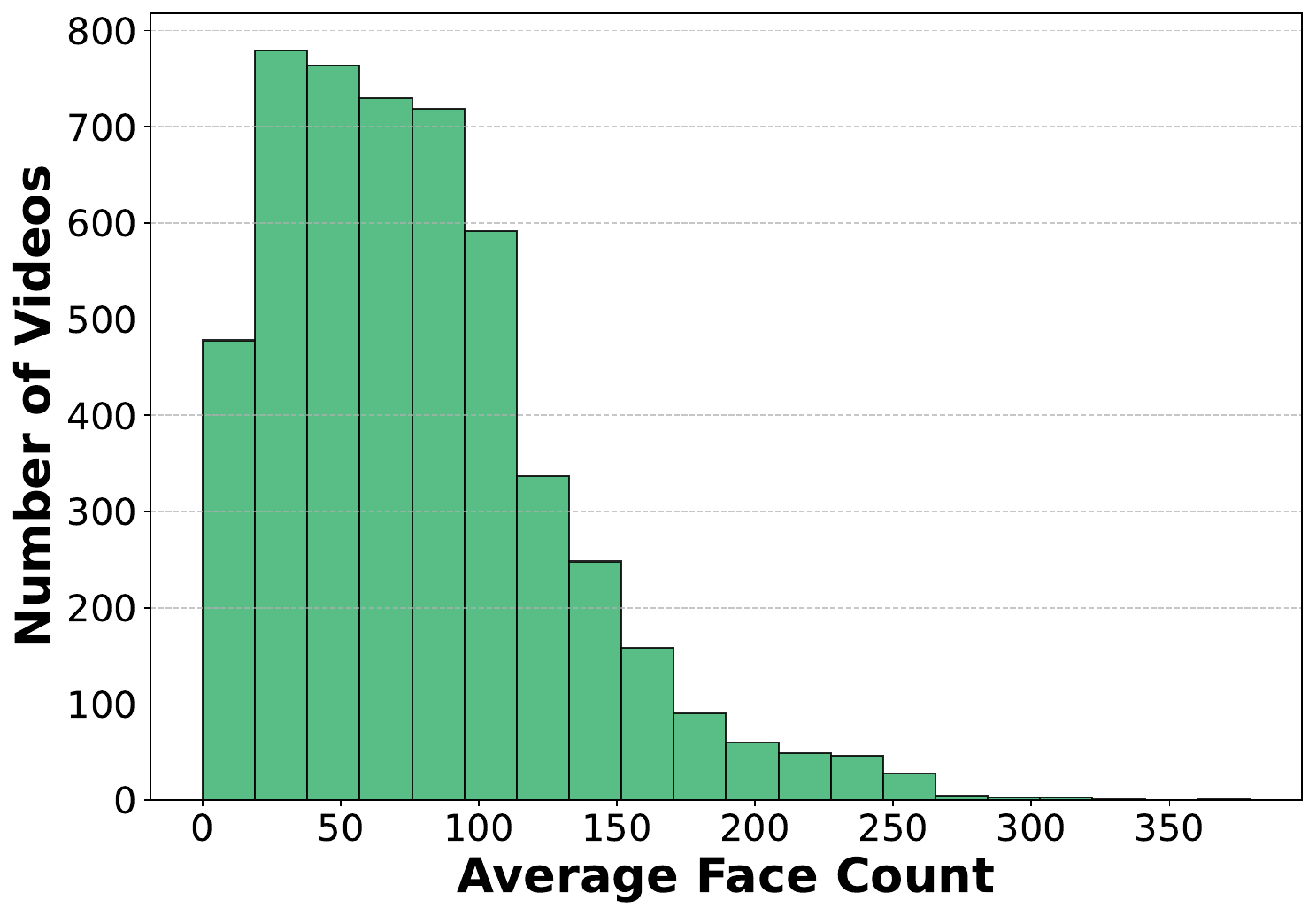}
\caption{Overview of data statistics. \textit{Left.} Ethnicity distribution across the dataset. \textit{Middle.} Gender distribution across the dataset. \textit{Right.} Face count distribution across the dataset. Abbreviations: Middle-E = Middle Eastern, Latino-H = Latino Hispanic, W = Women, M = Men.%using~\href{https://github.com/vladmandic/face-api?tab=readme-ov-file}{Face-api}. 
\textit{Right.} Per image detected face distribution.}
\label{fig:data_stat_fig}
\vspace{-3mm}
\end{figure*}

\subsubsection{Data Segmentation}
All videos were split into 5-second clips using FFmpeg~\cite{fmpeg2000}. This duration matches prior findings that expressive behaviours typically span 4–5 seconds before transitioning~\cite{ekman2003emotions,sharma2021audio}. To maximise diversity, we retained up to 35 segments per source video, covering the majority of group members despite dynamic framing. This yielded 5130 clips before preprocessing. %not mentioning #raw videos anymore to avoid any confusion.

\subsubsection{Data Preprocessing} 
All segmented clips underwent a manual verification process. Clips were excluded if they lacked a discernible group structure, such as the presence of a complete or partial group, or if the faces of all visible group members were not clearly detectable, which could compromise annotation accuracy. Additional filters removed low-resolution clips and those with minimal temporal information across frames. After this rigorous curation, 5091 high-quality clips remained. Each clip was standardised to 25 fps and 720p to give annotators a consistent viewing experience. %After quality filtering (discarding <1\% of segments due to annotation conflicts or low quality), 5091 clips remain.

\subsubsection{Data Annotation}
We used Labelbox~\cite{labelbox2024} for annotation (Interface: Fig.~\ref{fig:data}(c), Sample annotations: Fig.~\ref{fig:data}(d)). Clips were placed into 100 bins and assigned to 108 annotators (60 male, 48 female; average age 28 $\pm$ 5 years). Bins balanced workload and mitigated ordering bias. Three annotators viewed each clip and labeled group emotion (positive, neutral, negative), discrete emotion (happy, sad, fear, anger, neutral), intensity (high, medium, low), interaction type (cooperative, hostile, neutral) and action cues (e.g.\ smiling, clapping, shouting, dancing, singing, fighting, conversation, heated debate, protest, team activity). The annotators also verify the metadata generated by VLLM in previous phases. Each clip was labelled by three independent annotators, and the final label was decided by majority vote. If the three disagreed (i.e., no two votes matched), a fourth annotator was brought in to break the tie. Clips still lacking consensus were discarded. By requiring at least two matching votes and reporting high inter‐annotator agreement (as reported in Section~\ref{sec:agreement}), we ensure the final labels' reliability.

\subsection{Dataset Description}
Table~\ref{tab:datadetails} and Fig.~\ref{fig:data_stat_fig} summarise GAViD’s overall statistics and class distributions. Table~\ref{tab:sample_data} illustrates how we derive valence labels from the output probabilities of our three cross‐attention blocks (Visual-Audio, Visual-Context, Audio-Context). For each clip, we average these per‐class probabilities into a fused distribution and select the highest‐scoring class as the predicted valence. We apply the identical aggregation procedure to obtain discrete‐emotion labels. Fig.~\ref{fig:data_stat_fig} presents statistics on the average face count per video, as well as gender and ethnicity distributions, to provide fine-grained details of the proposed dataset.

\subsubsection{Dataset Statistics}
We split the 5091 clips in GAViD into non-overlapping train (3503), validation (542) and test (1046) sets, keeping all clips from the same source video in one split to avoid video-level bias. Each clip carries one of five discrete emotion labels (neutral, happy, sad, fear, anger). Following Plutchik’s emotion wheel~\cite{plutchik1980emotion}, the samples labelled `Excitement' and `Peaceful' were reclassified as `Happy,' and those labelled `Frustrated' were reclassified as `Anger.' The class-wise distribution reported in Table \ref{tab:datadetails} has more samples labelled neutral and happy. This imbalance is a common pattern observed in many benchmark datasets \cite{kosti2017emotion}.

\subsubsection{Inter-annotator Agreement}\label{sec:agreement} 
To assess the reliability of our annotations, we computed inter-annotator agreement using Cohen’s Kappa ($\kappa$) statistic~\cite{mchugh2012interrater}. We obtained high levels of agreement, with $\kappa=0.72$ for ternary valence labels and $\kappa=0.65$ for discrete emotion categories, supporting the quality and robustness of our dataset.

\subsubsection{Contextual Metadata} 
To enrich the dataset with high-level contextual understanding, each video is processed using the Video ChatGPT MLLM model~\cite{Maaz2023VideoChatGPT} with tailored prompts. The model extracts descriptive narratives for each clip, capturing contextual insights related to event type, environment and group interaction information. These contextual descriptions serve as metadata to enable context-aware GAR research.

\begin{table}[!t]
\centering
\renewcommand{\arraystretch}{1.35} % Increase row height (try 1.2–1.5)
\caption{Dataset Details. Here, `P': Positive, `N': Negative, `Ne': Neutral, `H': Happy, `S': Sad, `F': Fear, `A': Anger.}
\label{tab:datadetails}
\resizebox{.48\textwidth}{!}{%
    \begin{tabular}{|l|c|}
    \hline
    \textbf{Parameter}                & \textbf{Value} \\ \hline
    \#Samples (clips)                 & 5130  \\ \hline
    \#Corrupt samples                 & 39    \\ \hline
    \#Samples after filtering         & 5091 \\ \hline
    Average clip duration             & 5 sec \\ \hline
    Clip count per video              & 1–35  \\ \hline
    Dataset split                     & Train: 3503; Val: 542; Test: 1046  \\ \hline
    Affect label distribution         & P: 2600; N: 1189; Ne: 1302 \\ \hline
    Emotion label distribution        & Ne: 1522; H: 2428; A: 884; S: 201; F: 56 \\ \hline
    \end{tabular} 
} 
\end{table}

%\footnotetext{https://labelbox.com/}  \footnotemark[\value{footnote}] 
%====================================
%-*-*-*-*-*- Task Formulation & Methodology -*-*-*-*-*-
%====================================
\section{Proposed Methodology}\label{sec:method} 
This Section formulates the GAR task for the GAViD dataset and introduces the CAGNet GAR model (Fig.~\ref{fig:data_annotation}). 

\subsection{Task Formulation}
Let $\mathcal{G} = \{(\mathbf{V}_i, \mathbf{A}_i, \mathbf{C}_i, y^{(v)}_i, y^{(e)}_i)\}_{i=1}^N$ denote the GAViD dataset, where each 5-second clip $i$ contains visual features $\mathbf{V}_i$, audio features $\mathbf{A}_i$, scene-text features $\mathbf{C}_i$, a ternary valence label $y^{(v)}_i \in \{0,1,2\}$ and a five-class emotion label $y^{(e)}_i \in \{0,\dots,4\}$. The objective is to learn a predictive function $f_v$ mapping $(\mathbf{V},\mathbf{A},\mathbf{C})$ to $y^{(v)}$, by minimizing the empirical risk: $\mathcal{L} = \frac{1}{N} \sum_{i=1}^{N} \big[\, \ell(f_v(\mathbf{V}_i, \mathbf{A}_i, \mathbf{C}_i), y^{(v)}_i)]$, where $\ell$ denotes the cross-entropy loss.
%This encourages the predictors to leverage complementary cues, ensuring robustness and interpretability when any modality is absent.
% $\mathcal{L} = \frac{1}{N} \sum_{i=1}^{N} \big[\, \ell(f_v(\mathbf{V}_i, \mathbf{A}_i, \mathbf{C}_i), y^{(v)}_i) + \lambda\, \ell(f_e(\mathbf{V}_i, \mathbf{A}_i, \mathbf{C}_i), y^{(e)}_i) \,\big]$,
%Discrete-emotion term is commented out here and will be added in future extensions.

%Placed here for better formatting
\begin{figure*}[!t]
\centering
\includegraphics[width=1\linewidth, height=0.25\linewidth]{images/fig_CAGNet.png} 
\caption{Schematic architecture of CAGNet for Group Affect Recognition. Input video clips are processed into frames, audio and textual context, followed by preprocessing and modality-specific encoding. Encoded features are fused using a cross-attention mechanism and gated fusion. The fused representation is passed through an MLP to predict the group affect label. (b) shows the internal structure of the cross-attention block, including multihead attention, feed-forward layers and attention masking.}
\label{fig:data_annotation} 
\end{figure*}

%====================================
%-*-*-*-*-*- Baseline Models -*-*-*-*-*-
%====================================
\subsection{CAGNet}\label{sec:baseline}    
We propose CAGNet, a simple multimodal GAR model that fuses visual, audio and contextual information through the following phases.
 
\subsubsection{Modality-Specific Encoding}   
Each 5-second clip is decomposed into ten uniformly spaced 224$\times$224 RGB frames, a corresponding 16 kHz audio segment and its aligned transcript. These streams are embedded by frozen backbones: DINOv2~\cite{oquab2023dinov2} for vision, Wav2Vec 2.0~\cite{baevski2020wav2vec} for audio and XLM-RoBERTa~\cite{conneau2019unsupervised} for text. Mean pooling with padding masks yields fixed-length sequences $\mathbf{X}_{\!V}$, $\mathbf{X}_{\!A}$ and $\mathbf{X}_{\!C}$ in a common 768-D space while retaining the temporal order.
 
\subsubsection{Cross-Modal Alignment}  
As shown in Fig. \ref{fig:data_annotation}, every bimodal pair (vision audio, vision text, audio text) is processed by a masked multi-head cross-attention block (2 heads, 512-dim feed-forward). Queries originate from one stream and keys/values from the other, allowing each modality to query the other for cues it lacks. Padding masks silence missing time-steps, and residual connections preserve the original embeddings, producing three length-matched, cross-aligned representations.
 
\subsubsection{Gated Fusion}  
The cross-aligned embeddings are concatenated and passed through a squeeze–excitation gate inspired by~\cite{plosone,multiview}. Channel-wise importance weights
$
\mathbf{w}=\sigma\!\bigl(\mathrm{MLP}(\mathrm{AvgPool}([\![\cdot]\!]))\bigr)
$
re-scale the concatenation, yielding a context-adaptive fused vector $\mathbf{z}=\mathbf{w}\odot[\![\cdot]\!]$. This gate lets the model dynamically up‐weight, for example, the audio stream when faces are occluded or the textual stream when the visual scene is ambiguous.
 
\subsubsection{Classification}  
The fused vector $\mathbf{z}$ is passed through a lightweight two-layer multilayer perceptron (MLP) to produce the valence logits. First, we apply layer normalisation to stabilise the feature distribution. A non-linear GELU activation then models the complex decision boundaries, followed by dropout for regularisation. Finally, a linear projection produces three logits, which are softmaxed to yield valence probabilities. We use an MLP for its simplicity, efficiency, and proven effectiveness in multimodal fusion.

CAGNet offers two further benefits. First, it processes full clips end-to-end, avoiding unreliable face detection in dynamic group scenarios. Second, it robustly handles missing modalities by applying modality dropout during training by randomly zeroing one input stream per example. At test time, a zero vector is passed for the missing modality. The case of a missing modality in a trimodal setup is different from a bimodal setup. In bimodal, two modal channels are active. In the missing modality case, three channels remain active with one set to zero. CAGNet robustly adapts to this case. %To handle the missing modality, we train with cross-entropy loss and apply \emph{modality dropout}, randomly zeroing one stream so the classifier remains effective when a modality is noisy or absent at test time.

%====================================
%-*-*-*-*-*- Results -*-*-*-*-*-
%====================================
\section{Experimental Results}\label{sec:results} 

% \subsection{Experimental Details}
\subsection{Implementation Details}
Experiments are performed on an NVIDIA RTX A5000 GPU. We froze pretrained backbones (DINOv2, Wav2Vec 2.0, XLM-RoBERTa) and trained only the cross-attention blocks, squeeze-excitation gate and MLP head. A dropout rate of 0.4 and a batch size of 16 are used. The models are trained for up to 50 epochs with AdamW (LR $1\times10^{-4}$, weight decay $10^{-4}$), categorical cross-entropy to valence outputs, and early stopping with a patience of 5 to avoid overfitting. 

\subsection{Models}
Alongside the proposed CAGNet, we also designed two baseline models to systematically explore alternative fusion strategies and assess the impact of contextual information and modality combinations. \textit{Baseline 1 (Merged Fusion)} is a simple fusion approach: it first obtains cross-attention embeddings for each modality pair, concatenates these representations and passes the result through an MLP classifier, without applying any gated attention. \textit{Baseline 2 (Hierarchical Fusion)} adopts a hierarchical cross-attention strategy: it fuses vision and audio streams via a cross-attention block to obtain a joint representation, then combines it with contextual features using a second cross-attention block, followed by an MLP for final classification. In contrast, as detailed in Section \ref{sec:baseline}, \textit{CAGNet} integrates both cross-attention and gated fusion in a unified framework.

%Placed here for better formatting
\begin{figure*}[!t]
\centering
\includegraphics[width=0.75\linewidth, height=0.6\linewidth]{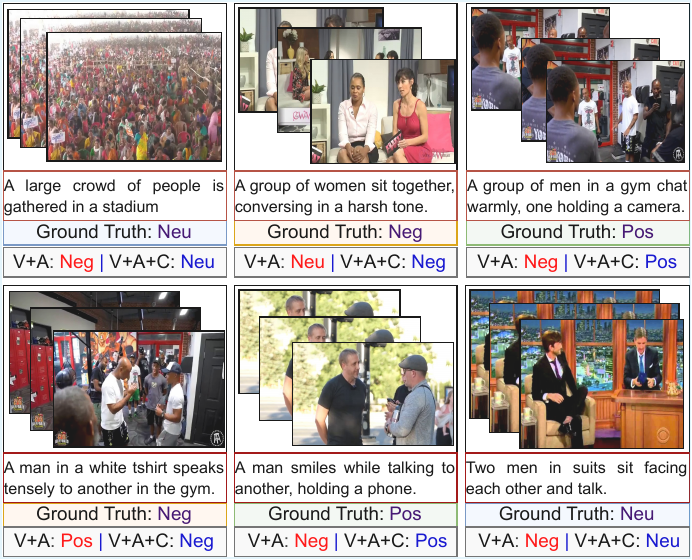}
\caption{Sample results for CAGNet on GAViD dataset under two settings: V+A and V+A+C. Pos: Positive, Neg: Negative and Neu: Neutral. V+A denotes training with audio-visual modalities, while V+A+C incorporates contextual information. The figure presents examples where adding context improves performance.}%These examples are randomly picked for illustration.
\label{fig:qualitative} 
\end{figure*}

\subsection{Results}
\vspace{-0.05in}
\subsubsection{Quantitative Analysis}
Table~\ref{tab:results} summarises the results and shows CAGNet outperforming both baselines, confirming the value of integrating cross-attention. The tri-modal setup (V+A+C) yields the strongest validation performance and maintains its lead on the test set, underscoring the importance of combining vision, audio and context. Examining bimodal variants reveals that visual+audio gives the highest accuracy, while visual+context shows the best F1 and audio+context trails behind, indicating that vision is the most informative modality in our clips. Notably, recent video-LLM-based state-of-the-art methods, i.e., Video-GPT and LLaVA-NeXT, achieve 51.92\%/0.525 and 54.82\%/0.388 (test Acc./F1), respectively, lagging behind even our simple fusion baselines and far below the tri-modal CAGNet, suggesting that generic video-language models without explicit context fusion are suboptimal for GAR. In a preliminary empirical analysis, we then simulated missing-modality conditions by zeroing out each input stream at test time and observed only modest performance drops, indicating that CAGNet handles absent modalities gracefully. Finally, retraining CAGNet on the combined train+validation splits further boosts test performance (to 66.21\% Acc. and 0.647 F1), demonstrating that more data enhances model robustness.
 %Finally, our modality‐dropout strategy ensures graceful degradation when inputs are missing, making CAGNet well-suited for in-the-wild scenarios.
% We evaluate all models using accuracy and F1 score to account for class imbalance.  

\begin{table*}[ht]
\centering
%\justify
%\justify
\renewcommand{\arraystretch}{1.3} % Increase row height (try 1.2–1.5)
\caption{Results for baseline, state-of-the-art models and CAGNet with various modality combinations (`V': Visual, `A': Audio, `C': Context). The results are reported on validation and test sets for models trained on the training split and combined training+validation splits. Here `+' indicates modality fusion, $\uparrow$ shows that higher values are better, and the highest values are marked in bold. }
\label{tab:results}
\resizebox{.99\textwidth}{!}{%
    \begin{tabular}{|l|c|c|c|c|c|c|c|}
    \hline
    \multirow{2}{*}{\textbf{Model}} & \multirow{2}{*}{\textbf{Modality}} 
      & \multicolumn{4}{c|}{\textbf{Train}} 
      & \multicolumn{2}{c|}{\textbf{Train+Val}} \\
    \cline{3-8}
    & & Val. Acc.\,$\uparrow$ & Val. F1\,$\uparrow$ 
      & Test Acc.\,$\uparrow$ & Test F1\,$\uparrow$ 
      & Test Acc.\,$\uparrow$ & Test F1\,$\uparrow$ \\
     \hline
   Video-GPT~\cite{zhuang2025video} &V+A&51.96\% &0.522&51.92\%&0.525&-&-     \\
     \hline
     LLaVA-NeXT~\cite{zhang2024llavanextvideo}&V+A    &54.14\%&0.379&54.82\%&0.388&-&-     \\
     \hline
    % available video model     &&&&&&&     \\

    \hline
    Baseline 1     & V + A + C & 62.62\% & 0.588 & 58.56\% & 0.554 & 61.55\% & 0.580 \\
    \hline
    Baseline 2     & V + A + C & 62.01\% & 0.573 & 57.47\% & 0.559 & 60.12\% & 0.588 \\
    \hline
    \multirow{4}{*}{\textbf{CAGNet}}
      & V + A     & 63.42\% & 0.588 & 59.53\% & 0.565 & 60.89\% & 0.597 \\
    \cline{2-8}
      & V + C     & 62.37\% & 0.595 & 59.13\% & 0.577 & 60.12\% & 0.606 \\
    \cline{2-8}
      & A + C     & 61.12\% & 0.584 & 58.29\% & 0.558 & 59.11\% & 0.591 \\
    \cline{2-8}
      & V + A + C & \textbf{64.81\%} & \textbf{0.609} & \textbf{63.20\%} & \textbf{0.614} & \textbf{66.21\%} & \textbf{0.647} \\
    \hline
    \end{tabular}%
}
\end{table*}

\begin{table}[!ht]
\centering
\renewcommand{\arraystretch}{1.3} % Increase row height (try 1.2–1.5)
\caption{Performance comparison of baseline methods using accuracy and F1-score as evaluation metrics for emotion recognition.}
\label{tab:overall_results}
\resizebox{0.49\textwidth}{!}{%
\begin{tabular}{|l|c|c|c|c|}
\hline
Model & \textbf{Val. Acc.}\,$\uparrow$ & \textbf{Val. F1}\,$\uparrow$ & \textbf{Test Acc.}\,$\uparrow$ & \textbf{Test F1}\,$\uparrow$ \\
\hline
Baseline 1 & 60.55\% & 0.434 & 58.33\% & 0.411 \\
\hline
Baseline 2 & 60.95\% & 0.410 & 60.01\% & 0.398 \\
\hline
\textbf{CAGNet}   & 63.55\% & 0.464 & 61.33\% & 0.458 \\
\hline
\end{tabular}%
}
\end{table}

\subsubsection{Qualitative Analysis}
To further demonstrate the effectiveness of incorporating contextual information, we present qualitative results on the proposed GAViD dataset. These examples highlight that while audio-visual cues (V+A) are sufficient in scenarios with clear signals such as dancing, singing, or protest activities, performance drops in cases where subtle facial expression changes dominate the interaction. In such instances, contextual cues provide complementary information, enabling more accurate predictions. For example, distinguishing between speaking in a harsh tone (negative) and chatting warmly (positive) often requires an understanding of the overall conversational context rather than relying solely on facial or audio features. As shown in Fig. \ref{fig:qualitative}, the inclusion of contextual information (V+A+C) helps the model capture these nuanced interactions more effectively, thereby improving prediction performance.

%====================================
%-*-*-*-*-*- Ethics -*-*-*-*-*-
%====================================
\section{Ethics, Licensing \& Reproducibility}\label{sec:ethics} 
This section details the ethical protocols, reproducibility measures and open access policies followed in constructing the GAViD dataset.
% \textcolor{blue}{Talk about:
% $\bullet$ licensing and access (maintenance also... Zenodo); 
% $\bullet$ reproducibility (\textbf{GitHub, Code released...} Reproducibility statement)
% $\bullet$ Ethical and privacy considerations (inc. Ethical protocols followed for compiling, sharing and processing the data, mention \href{https://support.google.com/youtube/answer/2797468}{https://support.google.com/youtube/answer/2797468}); 
% $\bullet$ supplementary materials;}

% \noindent \textcolor{blue}{
% $\bullet$ The paper should include a \textbf{link} to the dataset/website for review (Zenodo repo. \dots Github repo).
% $\bullet$ Any additional or \textbf{supplementary material} can be provided via a website about the dataset that is linked in the paper**.
% }

\subsection{Ethical and Privacy Considerations}
We prioritize the ethical handling of all GAViD content to uphold privacy, minimise risk and ensure responsible community use.

\begin{itemize} %[leftmargin=*]
\item \textbf{Data Collection and Consent:} The data collection and annotation strictly followed established ethical protocols in line with YouTube’s Terms, which state ``Public videos with a Creative Commons license may be reused"~\cite{youtube_tos,youtube_api_tos}. We downloaded only public-domain videos licensed under Creative Commons (CC BY 4.0), which ``allows others to share, copy and redistribute the material in any medium or format, and to adapt, remix, transform, and build upon it for any purpose, even commercially"~\cite{youtube_policy,cc_by}. %Specifically, as per YouTube's official statement, “If your video contains someone else’s personal information without their consent, it may be removed.” 

\item \textbf{Privacy:} All content was reviewed to ensure no private or sensitive information is present. Faces are included only from public domain videos as needed for group affect research; only group-level content is released, with no attempt or risk of individual identification. Other personally identifiable information, such as names, addresses and contacts, was removed.

\item \textbf{Ethical Review:} All data collection and release protocols were reviewed and approved by the affiliated institution's ethics board.

\item \textbf{Responsible and Fair Use:} Dataset documentation includes a fair-use statement. Users are advised not to deploy models trained on GAViD in real-world applications without safeguards. The dataset and models are for research and educational use and must not be used for commercial, surveillance, or profiling activities.
\end{itemize}

\subsection{Access, Licensing and Maintenance}
\begin{itemize} %[leftmargin=*]
\item \textbf{Access:} For long-term accessibility, the dataset is released on \href{https://zenodo.org/records/15448846}{\underline{https://zenodo.org/records/15448846}}, addressed as `Repository.'
%The GAViD dataset and the corresponding \textit{DOI} will be made publicly available after the paper's acceptance (currently not shared for double-blind anonymity).

% \href{https://zenodo.org/records/15448846}{\underline{GAViD repository}} (\href{https://zenodo.org/records/15448846}{\underline{zenodo.org/records/15448846}}, 
% DOI: \href{https://zenodo.org/records/15448846}{\underline{10.5281/zenodo.15448846}}).

\item \textbf{Licensing:} The GAViD dataset is released under the \emph{CC BY 4.0 license}, which allows unrestricted use, distribution, and reproduction in any medium, provided the original work is cited. Additionally, a custom license entry in Zenodo specifies that we prohibit the use of GAViD for surveillance, profiling, or automated decision-making without human oversight; models should remain within research or educational settings.
%The dataset will be released under a \emph{Custom License}. Access will be granted upon request and approval, in accordance with the Responsible Use policy. The license will prohibit use of GAViD for profit-making, surveillance, or profiling; models and data are strictly limited to research or educational purposes.

\item \textbf{Maintenance:} We will maintain version control and keep updating the documentation and future versions in the same \href{https://zenodo.org/records/15448846}{\underline{Repository}}.

%We will maintain version control and update future data versions and documentation. 

% \href{https://zenodo.org/records/15448846}{\underline{GAViD repository}}.
\end{itemize}

\subsection{Reproducibility and Documentation}
\begin{itemize} %[leftmargin=*]
\item \textbf{Reproducibility:} To promote transparency and further research in GAR and on the GAViD dataset, a GitHub repository containing code for dataset processing, label determination and baseline experiments is publicly released at \href{https://github.com/deepakkumar-iitr/GAViD}{\underline{https://github.com/deepakkumar-iitr/GAViD}}. 
%will be made publicly available after the paper's acceptance (currently not shared for double-blind anonymity). %\href{https://github.com/deepakkumar-iitr/GAViD}{\underline{https://github.com/deepakkumar-iitr/GAViD}}.
The repository includes comprehensive documentation on usage, a detailed \texttt{README} file, software dependencies and environment setup to support Reproducibility and future work.
\item \textbf{Supplementary Materials:} Additional details, such as dataset structure, file naming conventions, the full list of video download keywords, annotation guidelines, class-wise distributions for each data split and further results for discrete emotion recognition, are included in the \href{https://zenodo.org/records/15448846}{\underline{GAViD repository}}. %GitHub repository, which will be shared after the paper's acceptance (currently not shared for double-blind anonymity). 
\end{itemize}

\section{Conclusion}
This work contributes both a structured multimodal dataset and a modeling paradigm that supports quantitative analysis, representation, and understanding of collective affect in dynamic social systems, aligning with the broader goals of computational social systems research. GAViD dataset fills a critical gap by providing a large, richly annotated collection of real‐world group interactions with synchronised video, audio, and context labels. Its combination of valence, discrete emotion, and detailed contextual metadata makes it uniquely suited for training and evaluating models that must interpret nuanced group dynamics under uncontrolled conditions. By offering consistent annotations of emotion intensity, interaction types, and action cues alongside core affect labels, GAViD enables the development of systems that more accurately adapt to group moods in various settings, including human-computer interaction, customer analytics, public safety monitoring, and immersive entertainment. We believe GAViD’s comprehensive annotations will drive advances in context‐aware group affect recognition and enable applications that intelligently respond to collective emotional cues. 

We also propose CAGNet, a multimodal architecture integrating visual, audio and contextual features for group-level affect prediction, establishing a baseline for future research. GAViD’s rich multimodal and contextual annotations are expected to encourage research on valence and emotion analysis in GAR. In addition to valence classification, this paper also presents the results for discrete emotion recognition. We also plan to expand GAViD beyond group-level affect, incorporating frame-level and individual-level emotion annotations as well as increasing the dataset size. Further directions include exploring finer-grained temporal dynamics, domain adaptation for real-time deployment and transfer learning across related affective tasks. Such extensions will enable even more granular analysis and foster new research on group and individual affect in complex, real-world video settings.\vspace{-.1in}
%%%%%%%%%%%%%%%%%%%%%%%%%%%%%%%%%%%%%%%%%%%%%%

\vspace{2mm}
\section*{Acknowledgments}
This work was carried out at the Machine Intelligence Lab, Indian Institute of Technology Roorkee, Roorkee, Uttarakhand, India, during Dr. Deepak Kumar’s Ph.D. tenure.

\section*{Declaration of Generative AI Use}
The authors declare that a generative AI tool (e.g., ChatGPT) was used exclusively for language editing, including paraphrasing and grammatical refinement, during the preparation of this manuscript. All concepts, analyses, and results presented herein are the original work of the authors and were developed under their direct supervision and full responsibility.

\bibliography{ref}

% \bibliographystyle{IEEEtran}

% \bibitem{ref1}
% {\it{Mathematics Into Type}}. American Mathematical Society. [Online]. Available: https://www.ams.org/arc/styleguide/mit-2.pdf

% \bibitem{ref2}
% T. W. Chaundy, P. R. Barrett and C. Batey, {\it{The Printing of Mathematics}}. London, U.K., Oxford Univ. Press, 1954.

% \bibitem{ref3}
% F. Mittelbach and M. Goossens, {\it{The \LaTeX Companion}}, 2nd ed. Boston, MA, USA: Pearson, 2004.

% \bibitem{ref4}
% G. Gr\"atzer, {\it{More Math Into LaTeX}}, New York, NY, USA: Springer, 2007.

% \bibitem{ref5}M. Letourneau and J. W. Sharp, {\it{AMS-StyleGuide-online.pdf,}} American Mathematical Society, Providence, RI, USA, [Online]. Available: http://www.ams.org/arc/styleguide/index.html

% \bibitem{ref6}
% H. Sira-Ramirez, ``On the sliding mode control of nonlinear systems,'' \textit{Syst. Control Lett.}, vol. 19, pp. 303--312, 1992.

% \bibitem{ref7}
% A. Levant, ``Exact differentiation of signals with unbounded higher derivatives,''  in \textit{Proc. 45th IEEE Conf. Decis.
% Control}, San Diego, CA, USA, 2006, pp. 5585--5590. DOI: 10.1109/CDC.2006.377165.

% \bibitem{ref8}
% M. Fliess, C. Join, and H. Sira-Ramirez, ``Non-linear estimation is easy,'' \textit{Int. J. Model., Ident. Control}, vol. 4, no. 1, pp. 12--27, 2008.

% \bibitem{ref9}
% R. Ortega, A. Astolfi, G. Bastin, and H. Rodriguez, ``Stabilization of food-chain systems using a port-controlled Hamiltonian description,'' in \textit{Proc. Amer. Control Conf.}, Chicago, IL, USA,
% 2000, pp. 2245--2249.

% \end{thebibliography}

\newpage

\section*{Biography Section}
 

% \end{IEEEbiography}


\vspace{-0.3in}
\begin{IEEEbiography}[{\includegraphics[width=1in,height=1.25in,clip,keepaspectratio]{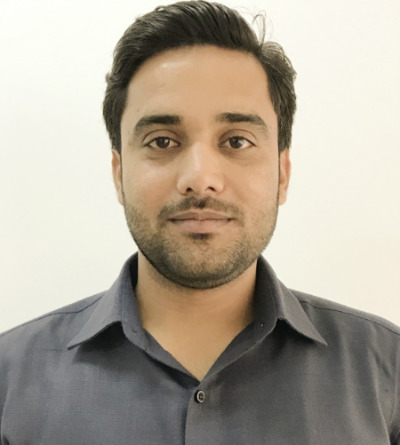}}]{Deepak Kumar} received his B.Tech. and M.Tech. degrees in Computer Science and Engineering in 2012 and 2016, respectively, and his Ph.D. from the Indian Institute of Technology Roorkee, India, in 2025. His research interests include affective computing, human–computer interaction, multimodal and interpretable artificial intelligence, and deepfake detection. He has published his work in leading journals and conferences. More information is available on his webpage: https://deepakkumar-iitr.github.io/
\end{IEEEbiography}

\vspace{-0.3in}
\begin{IEEEbiography}[{\includegraphics[width=1in,height=1.25in,clip,keepaspectratio]{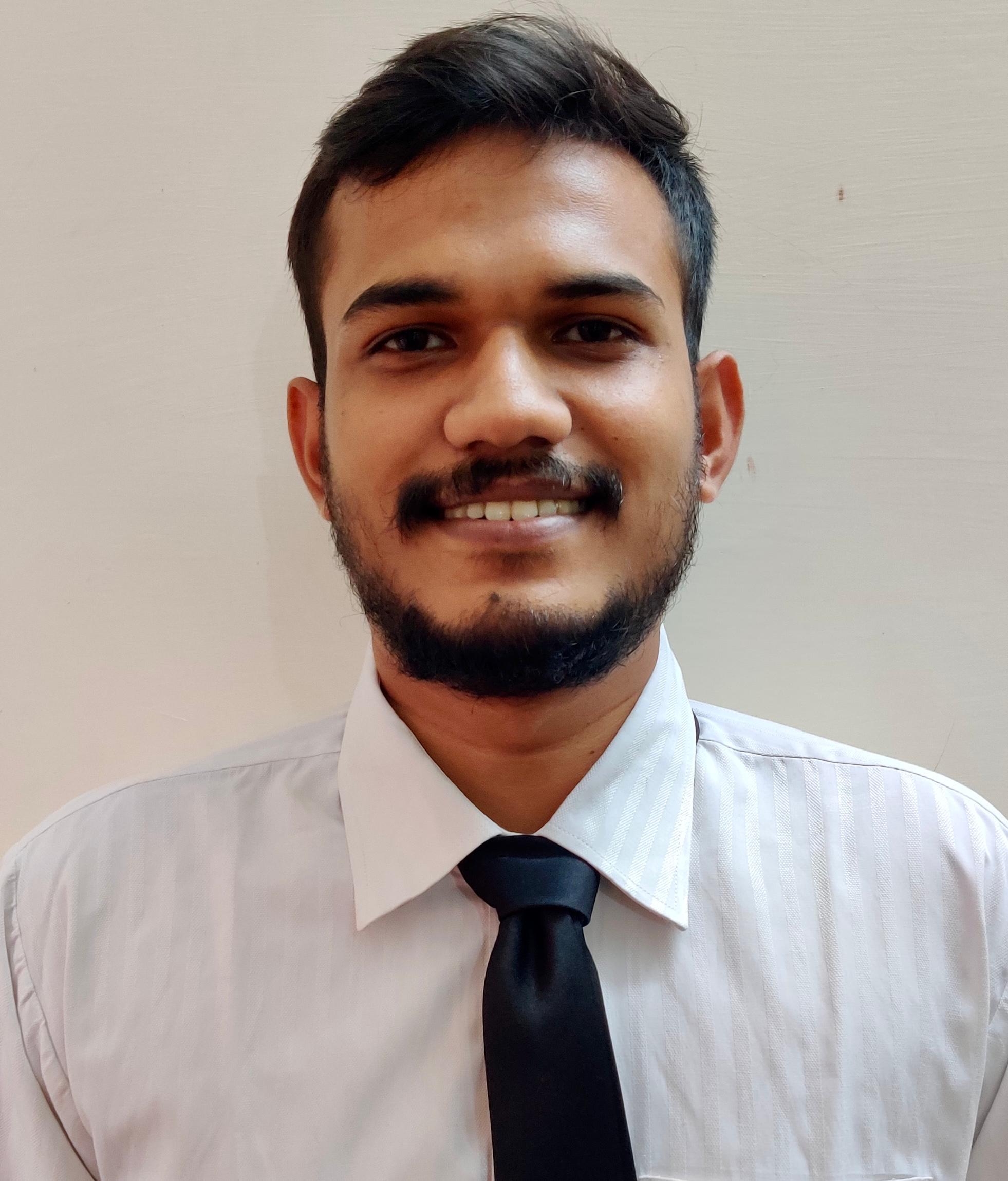}}]{Abhishek Pratap Singh} received his Integrated Post Graduate (B.Tech+M.Tech) degree in Information Technology in 2023 from IIIT Gwalior. He is a Ph.D student at IIT Roorkee, India. He works on Affective Computing, Multimodal \& Interpretable AI. For more information, visit his webpage https://aps19.github.io/.
\end{IEEEbiography}

\vspace{-0.3in}
\begin{IEEEbiography}[{\includegraphics[width=1in,height=1.25in,clip,keepaspectratio]{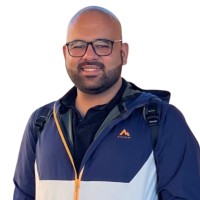}}]{Puneet Kumar} (Member, IEEE) received his B.E. and M.E. degrees in Computer Science in 2014 and 2018, respectively, and his Ph.D. from the IIT Roorkee, India in 2022. He is an Assistant Professor at the Indian Institute of Technology Ropar, India and an Affiliated Scientist at the Meditation Research Program, USA.
He works on Affective Computing, Multimodal \& Interpretable AI, Cognitive Neuroscience and Mental Health; has published in top journals \&
conferences and received institute medal in M.E., best Ph.D. thesis award and several best paper awards. For more information, visit his webpage www.puneetkumar.com.
\end{IEEEbiography}

\vspace{-0.3in}
\begin{IEEEbiography}[{\includegraphics[width=1in,height=1.25in,clip,keepaspectratio]{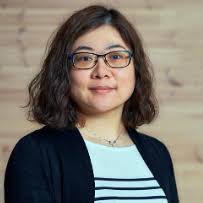}}]{Xiaobai Li} (Senior Member, IEEE) received her
B.Sc. and M.Sc. in 2004 and 2007, and her Ph.D. from the University of Oulu in 2017. She is a ZJU100 Professor at the State Key Lab of Blockchain and Data Security, Zhejiang University, and an Adjunct Professor at the University of
Oulu. Her research covers Affective Computing, Facial and Micro-Expression Recognition, and Remote Physiological Signal Measurement. She has co-chaired international workshops at CVPR, ICCV, FG, and ACM MM, and is an Associate
Editor for IEEE TMM, TCSVT, Frontiers in Psychology, and Image and Vision Computing. For details, visit her webpage xiaobaili-uhai.github.io/.
\end{IEEEbiography}

\vspace{-0.3in}
\begin{IEEEbiography}[{\includegraphics[width=1in,height=1.25in,clip,keepaspectratio]{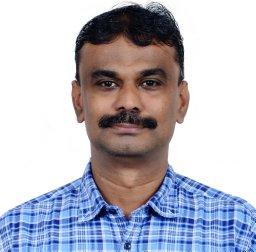}}]{Balasubramanian Raman} (Senior Member, IEEE) received his B.Sc. and M.Sc. degrees from the University of Madras in 1994 and 1996, respectively, and his Ph.D. from the IIT Madras in 2001. He is the Head and Chair Professor in the Department of Computer Science and a Joint
Faculty in the Mehta Family School of Data Science \& AI at IIT Roorkee, India. He has published over 200 research papers in reputed journals and conferences. His research interests include Machine Learning, Image and Video Processing,
Medical Imaging, Computer Vision, and Pattern Recognition. For more
information, visit his webpage at http://faculty.iitr.ac.in/cs/bala.
\end{IEEEbiography}

\vfill

\end{document}